\documentclass[sigconf]{acmart}

\usepackage{graphicx} % Required for inserting images
\usepackage{booktabs} % For high-quality tables
\usepackage{xspace}
\usepackage{multirow}
\usepackage{array}
\usepackage{xcolor}
\usepackage{colortbl}
\usepackage{makecell}
\usepackage[most]{tcolorbox}

\AtBeginDocument{%
  }

\setcopyright{acmlicensed}
\copyrightyear{2026}
\acmYear{2026}
\acmDOI{XXXXXXX.XXXXXXX}
\acmConference[ICAIL '26]{International Conference on Artificial
  Intelligence and Law}{June 8--12, 2026}{Singapore, Singapore}
\acmISBN{978-1-4503-XXXX-X/2026/06}

\begin{document}

%%
%% The "title" command has an optional parameter,
%% allowing the author to define a "short title" to be used in page headers.

\title{Magis-Bench: Evaluating LLMs on Magistrate-Level Legal Tasks}

%%
%% The "author" command and its associated commands are used to define
%% the authors and their affiliations.
%% Of note is the shared affiliation of the first two authors, and the
%% "authornote" and "authornotemark" commands
%% used to denote shared contribution to the research.

\settopmatter{authorsperrow=3}

\author{Ramon Pires}
\orcid{0000-0002-0023-1971}
% \authornotemark[1]
\affiliation{%
  \institution{Maritaca AI}
  \city{Campinas}
  \state{São Paulo}
  \country{Brazil}
}
\email{ramon@maritaca.ai}

\author{Thales Sales Almeida}
\orcid{0009-0006-9568-9331}
\affiliation{%
  \institution{Maritaca AI}
  \city{Campinas}
  \state{São Paulo}
  \country{Brazil}
}
% \email{thales@maritaca.ai}

\author{Celio Larcher Junior}
\orcid{0000-0002-4861-6571}
\affiliation{%
  \institution{Jusbrasil}
  \city{Salvador}
  \state{Bahia}
  \country{Brazil}
}
% \email{celio.larcher@jusbrasil.com.br}

\author{Giovana Bonás}
\orcid{0009-0001-9460-8353}
\affiliation{%
  \institution{Maritaca AI}
  \city{Campinas}
  \state{São Paulo}
  \country{Brazil}
}
% \email{giovana@maritaca.ai}

\author{Hugo Abonizio}
\orcid{0000-0001-5208-0290}
\affiliation{%
  \institution{Maritaca AI}
  \city{Campinas}
  \state{São Paulo}
  \country{Brazil}
}
% \email{hugo@maritaca.ai}

\author{Marcos Piau}
\orcid{0009-0001-1490-3476}
\affiliation{%
  \institution{Jusbrasil}
  \city{Salvador}
  \state{Bahia}
  \country{Brazil}
}
% \email{marcos.piau@jusbrasil.com.br}

\author{Roseval Malaquias Junior}
\orcid{0000-0002-6005-0515}
\affiliation{%
  \institution{Maritaca AI}
  \city{Campinas}
  \state{São Paulo}
  \country{Brazil}
}
% \email{roseval@maritaca.ai}

\author{Thiago Laitz}
\orcid{0000-0001-7205-2094}
\affiliation{%
  \institution{Maritaca AI}
  \city{Campinas}
  \state{São Paulo}
  \country{Brazil}
}
% \email{thiago@maritaca.ai}

\author{Rodrigo Nogueira}
\orcid{0000-0002-2600-6035}
\affiliation{%
  \institution{Maritaca AI}
  \city{Campinas}
  \state{São Paulo}
  \country{Brazil}
}
% \email{rodrigo@maritaca.ai}

%%
%% By default, the full list of authors will be used in the page
%% headers. Often, this list is too long, and will overlap
%% other information printed in the page headers. This command allows
%% the author to define a more concise list
%% of authors' names for this purpose.
\renewcommand{\shortauthors}{Pires et al.}

% Abstract
\begin{abstract}

% Evaluating legal reasoning at scale requires more than measuring whether models can produce fluent arguments: in high-stakes settings, society depends on decision-makers’ ability to judge arguments—assessing doctrinal fit, evidentiary implications, and the sufficiency of justification under explicit standards. Yet benchmarks that stress this judgment-centric dimension of judicial work remain scarce, in part because open-ended answers are costly to grade reliably.

% Evaluating legal writing is challenging not only because responses are open-ended, but because they must be judged: high-stakes legal work depends on reliably assessing the strength of arguments, the applicability of doctrine, and the adequacy of reasoning under explicit standards. Yet scalable evaluation of this "argument judgment" capability remains limited, since expert grading is costly and general-purpose metrics fail to capture doctrinal and procedural requirements.

% Evaluating open-ended judicial writing remains difficult: responses must be legally correct, procedurally well-structured, and aligned with domain-specific grading standards, yet expert human assessment is costly and hard to scale.

% While much attention has been devoted to evaluating whether large language models can produce legal arguments, comparatively little work has examined their capacity to \emph{judge} such arguments---a function that is arguably as fundamental to a well-functioning legal system as advocacy itself.
Existing benchmarks for legal AI focus primarily on tasks where LLMs must produce legal arguments or documents, yet the capacity to \emph{judge} such arguments---weighing competing claims, applying doctrine to facts, and rendering reasoned decisions---is arguably as fundamental to a well-functioning legal system as advocacy itself. 
We introduce Magis-Bench, a benchmark for evaluating LLMs on magistrate-level writing tasks derived from recent Brazilian competitive examinations for judicial positions. Magis-Bench comprises 74 questions from eight examinations conducted between 2023 and 2025, including discursive legal analysis questions with multi-turn structure and practical exercises requiring the composition of complete civil and criminal judicial sentences. %Each question is accompanied by official evaluation rubrics that specify the expected legal concepts, analytical steps, and structural elements.
We evaluate 23 state-of-the-art LLMs using an LLM-as-a-judge methodology with four independent frontier models as evaluators. Our results show strong inter-judge agreement (Kendall's $W = 0.984$; pairwise Kendall's $\tau \ge 0.897$), with Google's Gemini-3-Pro-Preview achieving the highest average score (6.97/10), followed by Gemini-3-Flash-Preview (6.67) and Claude-4.5-Opus (6.46). Even the best-performing models score below 70\% of the maximum, indicating that judicial-level legal reasoning and writing remain challenging for current LLMs. We release the complete benchmark, model outputs, and evaluation code\footnote{\url{https://github.com/maritaca-ai/magis-bench}} to support further research on legal AI capabilities.
\end{abstract}

%%
%% The code below is generated by the tool at http://dl.acm.org/ccs.cfm.
%% Please copy and paste the code instead of the example below.
%%


%%
%% Keywords. The author(s) should pick words that accurately describe
%% the work being presented. Separate the keywords with commas.

\begin{teaserfigure}
  \includegraphics[width=\textwidth]{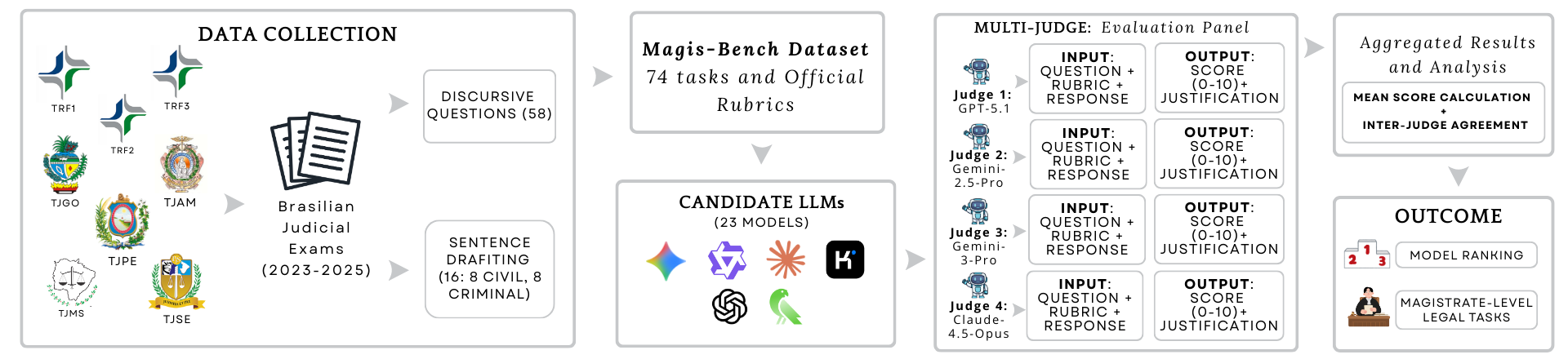}
  \caption{Overview of Magis-Bench: dataset construction and multi-LLM judging pipeline.}
  \Description[Magis-Bench workflow diagram]{Left-to-right flowchart of the benchmark. Data collection from Brazilian judicial exams (2023--2025) yields 58 discursive questions and 16 sentence-drafting tasks (8 civil, 8 criminal). These form the Magis-Bench dataset with 74 tasks and official rubrics. Twenty-three candidate LLMs are evaluated by a multi-judge panel of four models (GPT-5.1, Gemini-2.5-Pro, Gemini-3-Pro, and Claude-4.5-Opus). Each judge receives the question, rubric, and model response and outputs a 0--10 score with a justification. Scores are aggregated to compute mean scores and inter-judge agreement, producing a model ranking for magistrate-level legal tasks.}
  \label{fig:teaser}
\end{teaserfigure}

%%
%% This command processes the author and affiliation and title
%% information and builds the first part of the formatted document.
\maketitle

% Introduction
\section{Introduction}
\label{sec:introduction}

The growing adoption of LLMs in professional and specialized domains has created an urgent need for evaluation frameworks that can rigorously assess their capabilities beyond general-purpose tasks. %~\cite{healthbench, llmevalmed, xie2024finben, matlin2025financial, pires2026automated, chlapanis2025greekbarbench, xu2025claw}.
Existing open-ended benchmarks in the legal domain focus primarily on tasks where LLMs must produce legal arguments or documents~\cite{pires2026automated, chlapanis2025greekbarbench, shi2026plawbench, fan2025lexam}. Yet in a well-functioning legal system, the capacity to \emph{judge} arguments %---weighing competing claims, applying doctrine to facts, and rendering reasoned decisions---
is as fundamental as advocacy itself. While producing persuasive arguments tests one set of skills, judicial reasoning requires impartiality, comprehensive analysis of both sides, and authoritative resolution grounded in law. This distinction matters as LLMs are increasingly considered for applications involving adjudicative reasoning, from legal research assistance to decision support systems.

We introduce \textbf{Magis-Bench}, a benchmark for evaluating LLMs on magistrate-level legal tasks derived from recent Brazilian competitive examinations for judicial positions. In Brazil, judicial positions are filled through highly competitive public examinations that assess candidates' readiness to serve as judges. Magis-Bench comprises 74 questions from eight such examinations between 2023 and 2025, including discursive legal analysis questions with multi-turn structure and practical exercises requiring the composition of complete civil and criminal judicial sentences. Each question is accompanied by official evaluation rubrics that specify the expected legal concepts, analytical steps, and structural elements.

We evaluate 23 state-of-the-art LLMs using an LLM-as-a-judge methodology with four independent frontier models as evaluators. Our results show strong inter-judge agreement (Kendall's $W = 0.984$), with Google's Gemini-3-Pro-Preview achieving the highest average score (6.97/10). %, followed by Gemini-3-Flash-Preview (6.67) and Claude-4.5-Opus (6.46).
Even the best-performing models score below 70\% of the maximum, indicating that magistrate-level legal reasoning and writing remain challenging for current LLMs.

Our contributions are: (1) Magis-Bench, a benchmark of 74 magistrate-level legal tasks with official rubrics; (2) a multi-judge evaluation methodology that achieves high inter-judge agreement; and (3) a comprehensive evaluation of 23 LLMs, establishing performance baselines for judicial writing tasks. We release the benchmark, model outputs, and evaluation code to support further research.

The remainder of this paper is organized as follows. Section~\ref{sec:related_work} reviews related work. Section~\ref{sec:methodology} describes Magis-Bench and our evaluation methodology. Section~\ref{sec:results} reports experimental results. Section~\ref{sec:limitations} discusses limitations, and Section~\ref{sec:conclusion} concludes the paper.

\section{Related Work}
\label{sec:related_work}

% \paragraph{LLMs in Advocacy Roles.} 
Recent benchmarks assess LLMs' capacity to perform advocacy-oriented legal tasks, such as drafting legal documents and providing legal advice. 
% OAB-Bench~\cite{pires2026automated} requires models to generate complete legal essays %, with the best-performing model (Claude 3.5 Sonnet) passing thresholds across simulated exams.
% evaluated via LLM-as-a-judge using detailed checklists aligned with official grading criteria. 
% Rabula~\cite{pacheco2025rabula} combines multiple-choice questions with four generative tasks (document selection, brief drafting, and case resolution), reporting the inter-rater agreement between LLM evaluators and human experts. % evaluated via LLM-as-a-judge using detailed checklists aligned with official grading criteria.
OAB-Bench \cite{pires2026automated} and Rabula \cite{pacheco2025rabula} evaluate LLMs on the Brazilian Bar Examination (OAB), which tests advocacy competencies through essay tasks, using official FGV examination rubrics with LLM-as-a-judge evaluation. OAB-Bench achieves strong correlation with human expert scores.
LEXam \cite{fan2025lexam} provides a benchmark derived from 340 law school examinations across 116 courses at the University of Zurich, comprising questions in English and German.
% Beyond local bar exams, PLawBench~\cite{shi2026plawbench} models realistic legal workflows across 13 practice scenarios (client consultations, case analysis, document generation), encompassing 850 questions. %with ~12,500 binary evaluation criteria. % Notably, no model achieved strong performance on PLawBench, revealing significant gaps in deep legal reasoning and document structuring capabilities.
PLawBench~\cite{shi2026plawbench} evaluates practical legal skills in the Chinese domain through 850 questions across 13 scenarios. 
Other examination-based benchmarks include GreekBarBench~\cite{chlapanis2025greekbarbench} for the Greek legal system, KCL \cite{oh2025korean} for Korean canonical legal reasoning with bar exam questions and instance-level rubrics.

% \paragraph{LLMs in Judicial Decision-Making.}
A distinct research direction investigates whether LLMs can replicate judicial reasoning and produce sentencing decisions comparable to human judges. Posner \& Saran~\cite{posner2025judgeai} tested GPT-4 on a simulated appellate case involving war crimes, manipulating case framing and precedent alignment. The model exhibited formalist behavior, strictly following precedent while remaining insensitive to emotional appeals that influenced human judges. %---suggesting LLMs may lack the value-balancing intuition characteristic of experienced magistrates.
Gazal Ayal et al.~\cite{gazalayal2026sentencing} compared sentencing decisions from LLMs against 123 retired judges across two criminal cases. LLMs demonstrated substantially lower inter-model variability than human judges. %(e.g., Claude consistently assigned 2-year sentences with zero deviation vs. judges' mean of 2.14 years with SD=0.87), and their sentences aligned more closely with the judicial mean than individual judges' sentences. %This suggests LLMs can produce consistent, central-tendency decisions, though whether such uniformity is desirable remains debatable.
In the Chinese legal context, JuDGE~\cite{judge2024} benchmarks complete judgment document generation from factual case descriptions, demonstrating that retrieval-augmented approaches improve performance but substantial room for improvement remains.

% opção 1
% Magis-Bench specifically targets magistrate-level writing. This focus on judicial reasoning distinguishes our benchmark from advocacy-oriented evaluations and complements prior work on judicial decision simulation by providing a systematic, rubric-grounded assessment framework.

% opção 2
While existing benchmarks cover legal reasoning, judgment prediction, document generation, and bar examinations, none evaluates judicial competency using official certification criteria for judges. Magis-Bench fills this gap by using rubrics from competitive magistrate selection examinations in Brazil, testing whether LLMs can reason like judges as evaluated by professional examiners. %—distinct from replicating existing judicial documents or demonstrating advocacy skills.

% Methodology
\section{Methodology}
\label{sec:methodology}

This section describes how we constructed Magis-Bench and how we evaluate LLMs on it using rubric-grounded, multi-judge LLM-as-a-judge scoring.

\subsection{Data Collection}
\label{subsec:data_collection}

Magis-Bench is constructed from written examinations administered as part of competitive selection processes for substitute judges in Brazil. These examinations constitute the second phase of judicial selection, following the objective (multiple-choice) phase, and are designed to assess candidates' ability to apply legal knowledge in sophisticated writing tasks under examination conditions.

We collected examinations from eight judicial selection processes conducted between 2023 and 2025, comprising both federal (TRF1, TRF2, and TRF3) and state (TJMS, TJPE, TJGO, TJAM, and TJSE) courts. The selection criteria for including examinations were: (1) the examination must have been conducted in 2023 or later%, ensuring currency of legal content and examination standards
; (2) the written examination questions and official evaluation rubrics must be publicly available; and (3) the examination must include both discursive questions and practical sentence-drafting exercises. Table~\ref{tab:exams} summarizes the examinations included in the benchmark.

\begin{table}[ht]
\centering
\caption{Examinations included in Magis-Bench.}
\label{tab:exams}
\small
\begin{tabular}{llcc}
\hline
\textbf{Court} & \textbf{Organizer} & \textbf{Year} & \textbf{Questions} \\
\hline
TRF1 & FGV & 2023 & 6 \\
TRF2 & Internal Commission & 2024 & 14 \\
TRF3 & FGV & 2025 & 8 \\
TJMS & FGV & 2023 & 12 \\
TJPE & FGV & 2023 & 10 \\
TJGO & FGV & 2023 & 10 \\
TJAM & FGV & 2025 & 7 \\
TJSE & FGV & 2025 & 7 \\
\hline
\textbf{Total} & & & \textbf{74} \\
\hline
\end{tabular}
\end{table}

\subsection{Magis-Bench}
\label{sec:magis_bench}

Magis-Bench comprises 74 questions: 58 discursive questions and 16 sentence-drafting exercises (8 civil and 8 criminal). Each question is scored on a scale of 0 to 10 points based on official evaluation rubrics produced by the examination boards. The discursive questions present factual scenarios followed by one or more prompts requiring legal analysis; many are multi-turn (108 total turns across all discursive questions). For evaluation, we present each sub-question sequentially and allow the model to see its previous responses, simulating the examination context.

The sentence-drafting exercises require candidates to draft complete judicial decisions following Brazilian procedural law structure. %: \textit{relatório} (report), \textit{fundamentação} (reasoning), and \textit{dispositivo} (operative part).
Civil sentences involve disputes such as tax, administrative, and social security law, while criminal sentences require analysis of charges and defenses and, when conviction is warranted, sentencing guidelines. % (\textit{dosimetria da pena}).
Figure~\ref{fig:sentence_example} shows an example of a practical civil sentence question.

\begin{figure}[t]
\centering
% \fbox{\begin{minipage}[c][0.35\textheight][l]{0.95\linewidth}
\begin{tcolorbox}[
  width=0.95\linewidth,
  colback=black!2,        % fundo bem leve
  colframe=black!25,      % borda discreta
  boxrule=0.4pt,          % espessura da borda
  arc=2mm,                % cantos levemente arredondados
  left=3mm,right=3mm,top=2mm,bottom=2mm,
  title=\footnotesize\fontfamily{phv}\selectfont PRACTICAL CIVIL JUDGMENT EXAM,
  colbacktitle=black!6,
  coltitle=black!80,
  fonttitle=\bfseries,
]
%\centering
% \small
\footnotesize
%\fontfamily{phv}\selectfont
%\textbf{Placeholder: example of a practical sentence-drafting question}\\
%(Insert a cropped screenshot or a stylized template showing the prompt and required structure.)
% DEIDE COSTA propôs, em 23/07/2024, demanda indenizatória em face de TRANSPORTE KIKO LTDA. Pretende a autora a condenação da ré à indenização por danos materiais e compensação por danos morais. Sustenta, para tanto, que, em 20/07/2020, seu filho, DEIDINHO COSTA, ingressou no coletivo da ré, placa XYZ-1234, para retornar a sua residência. Durante o percurso, o motorista veio a colidir na traseira de outro coletivo, de placa PIU-7171. Embora socorrido, ele não sobreviveu às lesões. Pleiteia: \\
% (i) o ressarcimento do valor de R\$ 1.099,00, relativo aos gastos com os tratamentos subministrados na tentativa de recuperar seu filho e com o enterro; e \\
% (ii) o pagamento de indenização por danos morais. \\
% ... \\
% Com base na situação proposta no enunciado, que já vale como relatório (dispensada a repetição), \textbf{profira sentença enfrentando todos os pontos explícita e implicitamente abordados}. Ainda que entenda pelo acolhimento de alguma preliminar ou questão prejudicial, resolva todas as questões fáticas e de direito, de maneira fundamentada e estruturada nos termos do que determina o Código de Processo Civil. \\
% Valor: \textbf{10 pontos}, Máximo de \textbf{300 linhas}. \\

%\textbf{PRACTICAL CIVIL JUDGMENT EXAM}\\
%\\
XYZ Comércio Ltda., a business company operating in the retail trade sector, has tax-enforcement-registered debts in the amount of R\$ 200,000.00, relating to the contribution levied on payroll in favor of the National Commercial Apprenticeship Service (SENAC). The taxable events for such contributions occurred throughout the year 2016. Such assessments were never challenged either administratively or judicially.\\
Due to these debts, the company became subject to a tax enforcement proceeding (execução fiscal), filed by the Federal Government on 04/03/2017, to collect the said debt. The case was assigned to the 3rd Federal Tax Enforcement Court of the seat of the Judiciary Section.\\
...
\\
In its response to the objections, the Federal Government argued:\\
i) that it is a proper party to collect;\\
ii) being a proper party, the jurisdiction to process and adjudicate such collection by means of tax enforcement lies with the Federal Judiciary;\\
iii) the intercurring statute of limitations was not consummated;\\
iv) the contributions in favor of the ``S System'' levied on payroll were received by the 1988 Federal Constitution;\\
v) there is no limitation of 20 minimum wages on the tax base of the contributions to SENAC.\\
The records were submitted for judgment.\\
\textbf{In view of the data above (to which no facts created by the candidate must be added), \emph{render the judgment} (grounds/reasons and operative part), addressing each of the allegations with the appropriate legal basis and/or current prevailing understanding of the case law. Preparation of the report section is dispensed with.}
% \end{minipage}}
\end{tcolorbox}
\caption{Example of a practical sentence-drafting question in Magis-Bench: Civil sentence of TRF2.} %\textit{trf2\_comissao\_2024\_sentenca\_civel\_questao\_1}.}
\Description[Exam prompt excerpt]{Image containing the prompt text for the example practical civil sentence-drafting task.}
\label{fig:sentence_example}
\end{figure}

We evaluate a diverse set of LLMs on Magis-Bench, covering proprietary and open-source models across different sizes and architectures. The evaluated models include offerings from OpenAI, Anthropic, Google, Maritaca AI, and several open-source families.

Model outputs for most models were obtained through OpenRouter\footnote{\url{http://openrouter.ai/}}, except for OpenAI and Maritaca AI models, which were accessed via their respective APIs. We use each model's default temperature setting (temperature unspecified) rather than imposing a uniform value across all models. This approach ensures that each model operates under developer-recommended conditions, providing a fair assessment of real-world deployment performance. All other generation parameters use each model's default settings.

\subsection{Rubric-Grounded Multi-Judge Evaluation}
\label{sec:multi-judge}

Evaluating open-ended legal writing at scale is costly and difficult to standardize with human experts. We therefore employ an LLM-as-a-judge methodology \cite{mt_bench}, wherein strong LLMs evaluate generated responses against the official rubrics and assign a score from 0 to 10 with a brief justification.

For each response, the judge receives: (1) the original question (including the factual scenario and prompt); (2) the official evaluation rubric specifying expected elements and point allocation; and (3) the model's generated response. Rubric grounding helps reduce reliance on subjective preferences by anchoring evaluation to explicit, pre-defined criteria.

Single-judge evaluation can be sensitive to model-specific biases and inconsistent criterion weighting. To improve robustness, we use four independent frontier models as judges and report both per-judge scores and their aggregate (mean) across all questions.
Evaluation is blind: judges receive only the question, rubric, and response, without any identification of the candidate model. Together with rubric grounding, this mitigates stylistic preferences that could favor same-family candidates.

The four judge models used in our evaluation are:
\begin{itemize}
    \item \textbf{GPT-5.1} (OpenAI)
    \item \textbf{Gemini-2.5-Pro} (Google)
    \item \textbf{Gemini-3-Pro-Preview} (Google)
    \item \textbf{Claude-4.5-Opus} (Anthropic)
\end{itemize}

All judge evaluations are conducted with temperature set to 0 to maximize reproducibility and minimize variation in scoring.

To quantify consensus among judges, we report Kendall's $W$~\cite{kendalls_w} for overall concordance across all four judges and Kendall's $\tau$~\cite{kendalls_tau} for pairwise rank correlations, both computed over per-judge rankings of the evaluated models. $W$ ranges from 0 to 1 (higher indicates stronger consensus) and $\tau$ from $-1$ to $+1$.

% Experimental Results
\section{Results}
\label{sec:results}

This section presents the experimental results of evaluating 23 LLMs on Magis-Bench, including overall performance rankings, inter-judge agreement analysis, and examination of judge-specific evaluation patterns.

\subsection{Overall Performance}

Table~\ref{tab:results_ranked} presents the performance of all evaluated models across the four judges. Models are ranked by their average score across all judges, with scores representing the mean across all 74 questions in the benchmark (each scored 0--10).

The results reveal a clear performance hierarchy. Google's Gemini-3-Pro-Preview achieves the highest average score (6.97), followed by Gemini-3-Flash-Preview (6.67) and Claude-4.5-Opus (6.46). The top five positions are occupied exclusively by frontier models from Google, Anthropic, and OpenAI, with scores ranging from 6.18 to 6.97. Notably, even the best-performing model achieves less than 70\% of the maximum possible score, indicating that Magis-Bench presents a substantial challenge for current LLMs.

The mid-tier models (ranks 6--14) achieve scores between 4.06 and 5.55, including Claude-4.5-Sonnet, GPT-4.1, the Maritaca AI Sabiá models, and various reasoning-enhanced models. The lower tier (ranks 15--23) comprises smaller open-source models, primarily from the Qwen family, with scores ranging from 1.82 to 4.00.
Bootstrap 95\% confidence intervals (10{,}000 resamples over per-exam scores) confirm this tier structure: top-5 CIs overlap within the group but are separated from mid-tier models (e.g., rank 1 [6.34, 7.56] vs.\ rank 6 [4.89, 6.16]).
Averaged across models, discursive questions are the easiest (mean 4.68), followed by criminal sentence drafting (4.39) and civil sentence drafting (3.89).

\definecolor{pos3}{RGB}{34,139,34}    % dark green: +3 or more
\definecolor{pos2}{RGB}{102,194,102}  % medium green: +2
\definecolor{pos1}{RGB}{198,239,198}  % light green: +1
\definecolor{neg1}{RGB}{255,204,204}  % light red: -1
\definecolor{neg2}{RGB}{255,128,128}  % medium red: -2
\definecolor{neg3}{RGB}{220,60,60}    % dark red: -3 or less

% Macros for colored cells (diff = avg_rank - judge_rank, positive = judge ranked higher)
\newcommand{\pthree}[1]{\cellcolor{pos3}\textcolor{white}{#1}}
\newcommand{\ptwo}[1]{\cellcolor{pos2}{#1}}
\newcommand{\pone}[1]{\cellcolor{pos1}{#1}}
\newcommand{\nzero}[1]{#1}
\newcommand{\none}[1]{\cellcolor{neg1}{#1}}
\newcommand{\ntwo}[1]{\cellcolor{neg2}{#1}}
\newcommand{\nthree}[1]{\cellcolor{neg3}\textcolor{white}{#1}}

\begin{table*}[t]
    \centering
    \caption{Model rankings across four LLM judges. \colorbox{pos2}{Green}/\colorbox{neg2}{red} cells: judge ranked above/below mean (intensity to $\pm 3+$).}
    \label{tab:results_ranked}
    \small
    \setlength{\tabcolsep}{4pt}
    \renewcommand{\arraystretch}{0.9}
    \begin{tabular}{rlccccc}
        \toprule
        \textbf{Rank} & \textbf{Model} & \textbf{GPT-5.1} & \textbf{Gemini-2.5-Pro} & \textbf{Gemini-3-Pro} & \textbf{Claude-4.5-Opus} & \textbf{AVG} \\
        \midrule
        1  & Gemini-3-Pro-Preview          & \nzero{6.45} & \nzero{7.29} & \nzero{7.43} & \nzero{6.69} & 6.97 \\
        2  & Gemini-3-Flash-Preview        & \nzero{6.08} & \nzero{6.86} & \nzero{7.29} & \nzero{6.46} & 6.67 \\
        3  & Claude-4.5-Opus               & \none{5.89}  & \nzero{6.74} & \nzero{6.97} & \nzero{6.24} & 6.46 \\
        4  & GPT-5.1                       & \pone{6.06}  & \nzero{6.60} & \none{6.49}  & \none{5.75}  & 6.23 \\
        5  & Gemini-2.5-Pro                & \nzero{5.77} & \nzero{6.37} & \pone{6.67}  & \pone{5.91}  & 6.18 \\
        6  & Claude-4.5-Sonnet             & \nzero{5.30} & \nzero{5.86} & \none{5.63}  & \nzero{5.39} & 5.55 \\
        7  & GPT-4.1                       & \nzero{4.68} & \nzero{5.39} & \pone{5.73}  & \nzero{5.01} & 5.20 \\
        8  & Sabiá-4                       & \nzero{4.46} & \nzero{5.26} & \nzero{5.60} & \nzero{4.79} & 5.03 \\
        9  & Sabiá-3.1                     & \none{3.98}  & \nzero{4.77} & \none{4.76}  & \nzero{4.15} & 4.41 \\
        10 & DeepSeek-V3.2                 & \pone{4.04}  & \nzero{4.70} & \pone{4.90}  & \nthree{3.73}& 4.34 \\
        11 & Kimi-K2.5                     & \nzero{3.97} & \nzero{4.43} & \nthree{4.43}& \pone{3.83}  & 4.17 \\
        12 & Kimi-K2-Thinking              & \nzero{3.91} & \ntwo{4.26}  & \none{4.59}  & \pone{3.80}  & 4.14 \\
        13 & GPT-5-Mini                    & \none{3.62}  & \nzero{4.34} & \ptwo{4.66}  & \none{3.70}  & 4.08 \\
        14 & Sabiazinho-4                  & \pone{3.65}  & \ptwo{4.39}  & \nzero{4.49} & \none{3.70}  & 4.06 \\
        15 & Qwen3-235B-Thinking           & \none{3.48}  & \nzero{4.21} & \pthree{4.60}& \pthree{3.74}& 4.00 \\
        16 & Qwen3-235B-Instruct           & \none{3.40}  & \nzero{4.09} & \none{4.22}  & \nzero{3.64} & 3.84 \\
        17 & GPT-4.1-Mini                  & \ptwo{3.62}  & \nzero{4.06} & \pone{4.29}  & \nzero{3.38} & 3.84 \\
        18 & Sabiazinho-3                  & \nzero{3.11} & \none{3.13}  & \nzero{3.50} & \nzero{2.83} & 3.14 \\
        19 & Qwen3-30B-Instruct            & \nzero{2.59} & \pone{3.24}  & \nzero{3.28} & \nzero{2.51} & 2.91 \\
        20 & Qwen2.5-72B-Instruct          & \nzero{2.58} & \nzero{2.83} & \nzero{2.92} & \nzero{2.16} & 2.62 \\
        21 & Qwen3-30B-Thinking            & \nzero{2.33} & \nzero{2.55} & \nzero{2.50} & \nzero{2.01} & 2.35 \\
        22 & Qwen2.5-14B-Instruct          & \nzero{2.10} & \nzero{1.95} & \nzero{2.30} & \nzero{1.82} & 2.04 \\
        23 & Qwen3-8B                      & \nzero{1.86} & \nzero{1.81} & \nzero{2.13} & \nzero{1.46} & 1.82 \\
        \bottomrule
    \end{tabular}
\end{table*}

\subsection{Inter-Judge Agreement}

A central question in LLM-as-a-judge evaluation is whether different judge models produce consistent assessments. Our multi-judge methodology enables direct measurement of inter-judge agreement, providing insight into the robustness of the evaluation.

\paragraph{Overall Concordance.} The four judge models exhibit remarkably high agreement in their rankings. Kendall's coefficient of concordance reaches $W = 0.984$, indicating near-perfect consensus on the relative ordering of models. This level of agreement substantially exceeds what would be expected by chance and suggests that the judges are responding to genuine quality differences in model outputs.

\paragraph{Pairwise Correlations.} Table~\ref{tab:kendall_tau} presents Kendall's $\tau$ coefficients for all six pairwise combinations of judges. All correlations exceed 0.89, with a mean of $\tau = 0.913$. The highest agreement is observed between GPT-5.1 and Gemini-2.5-Pro ($\tau = 0.945$), while the lowest, though still strong, is between Gemini-3-Pro-Preview and both Gemini-2.5-Pro and Claude-4.5-Opus ($\tau = 0.897$). All p-values are below $10^{-13}$, confirming statistical significance.

\begin{table}[ht]
\centering
\caption{Pairwise Kendall's $\tau$ correlation between judge models.}
\label{tab:kendall_tau}
\setlength{\tabcolsep}{3pt}
\resizebox{\columnwidth}{!}{%
\begin{tabular}{lcccc}
\toprule
 & \textbf{GPT-5.1} & \textbf{Gemini-2.5-Pro} & \textbf{Gemini-3-Pro} & \textbf{Claude-4.5-Opus} \\
\midrule
GPT-5.1         & 1.000 & 0.945 & 0.905 & 0.913 \\
Gemini-2.5-Pro  & 0.945 & 1.000 & 0.897 & 0.921 \\
Gemini-3-Pro    & 0.905 & 0.897 & 1.000 & 0.897 \\
Claude-4.5-Opus & 0.913 & 0.921 & 0.897 & 1.000 \\
\bottomrule
\end{tabular}%
}
\end{table}

\subsection{Judge-Specific Patterns}

Despite the high overall agreement, the colored cells in Table~\ref{tab:results_ranked} reveal subtle but interpretable differences in how judges evaluate certain models.

\paragraph{Scoring Tendencies.} Gemini-3-Pro-Preview tends to assign relatively higher scores across the board, as reflected in its column showing the highest raw scores for most models. Conversely, Claude-4.5-Opus tends toward slightly more conservative scores, particularly for mid-tier models. GPT-5.1 and Gemini-2.5-Pro occupy intermediate positions.

\paragraph{Notable Divergences.} A few specific cases warrant attention. Claude-4.5-Opus assigns DeepSeek-V3.2 a notably lower score than other judges (rank divergence of $-3$), suggesting this judge may be more critical of certain response patterns exhibited by DeepSeek. Similarly, Gemini-3-Pro-Preview ranks Qwen3-235B-Thinking substantially higher than the consensus ($+3$). These divergences, while informative, do not substantially affect the overall ranking given the high concordance observed.

\paragraph{Judge Calibration.} To verify that judges can identify optimal performance, we generated responses for all 74 questions using GPT-5.2 (reasoning effort high) with access to official rubrics as privileged information. All four judges assigned near-perfect scores to these oracle responses (mean 9.957; individual scores $\geq 9.87$), substantially higher than the best-performing evaluated model (6.97), confirming that judges can distinguish truly excellent from typical outputs.

\paragraph{Implications.} The high inter-judge agreement validates the use of LLM-as-a-judge methodology for Magis-Bench evaluation. The consistency across four independent frontier models from three different providers suggests that the evaluation captures genuine quality differences rather than arbitrary preferences. The minor divergences observed provide additional insight into how different judges weight various aspects of legal writing, but do not undermine the reliability of the overall rankings.

\subsection{Robustness of the Judge Panel}
\label{sec:robustness}

A leave-one-judge-out analysis yields Kendall's $\tau \geq 0.976$ against the full ranking in all four runs, with no model moving more than two positions. A self-bias check, comparing each judge-candidate's self-score with the mean of the other three judges, finds differences from $-0.29$ (Claude-4.5-Opus) to $+0.62$ (Gemini-3-Pro-Preview), with mean $+0.09$. This reflects general leniency rather than self-favoritism: Gemini-3-Pro-Preview assigns the top score to 19 of 23 models.

% Limitations
\section{Limitations}
\label{sec:limitations}

Magis-Bench provides a rubric-grounded way to study magistrate-level writing, but it has important limitations. First, our main results rely on LLM-as-a-judge scoring (Section~\ref{sec:methodology}); although we mitigate model-specific biases by using four independent frontier judges and observe high inter-judge agreement, agreement is not the same as correctness, and judges may systematically reward surface features (e.g., verbosity or formality) or miss subtle doctrinal errors. A natural next step is targeted human validation on a stratified subset of questions and model outputs, both to calibrate absolute score levels and to audit whether judge explanations and scores track the official rubrics as human graders would apply them. Another limitation is that, because the rubrics reflect examination-board expectations, they may encode particular preferences that differ from real-world best practices; extending the benchmark to additional sources and task types would improve coverage.
Examinations span 2023--2025 (several administered after most models' training cutoffs) and rubrics are released separately from the questions, though inclusion in proprietary training corpora cannot be verified. The benchmark uses only publicly released materials and does not advocate deploying LLMs to adjudicate real cases.

% Conclusion
\section{Conclusion}
\label{sec:conclusion}

We introduced Magis-Bench, a benchmark for evaluating LLMs on magistrate-level legal tasks derived from recent Brazilian competitive examinations for judicial positions. The benchmark comprises 74 questions from eight examinations between 2023 and 2025. Each question is accompanied by official evaluation rubrics that enable systematic, criteria-grounded assessment.

Our evaluation of 23 state-of-the-art LLMs using a multi-judge methodology with four frontier models revealed several key findings. First, inter-judge agreement is remarkably high (Kendall's $W = 0.984$), and judges demonstrate appropriate calibration: when evaluating oracle responses generated with access to official rubrics, all four judges assigned near-perfect scores, confirming they can reliably distinguish truly excellent responses from typical model outputs. Second, even the best-performing models achieve less than 70\% of the maximum possible score, indicating that magistrate-level legal reasoning and writing remain challenging for current LLMs. Third, the performance hierarchy is largely consistent across judges, with Google's Gemini-3-Pro-Preview, Gemini-3-Flash-Preview, and Claude-4.5-Opus occupying the top positions.

Magis-Bench addresses a gap in legal AI evaluation by focusing on magistrate-level legal tasks rather than advocacy-oriented skills. This distinction is important as LLMs are increasingly considered for applications that involve adjudicative reasoning, from legal research assistance to decision support systems.

The average cost to evaluate a single model with a single judge ranged from \$4.43 to \$5.91 USD across the four judges, demonstrating the feasibility of large-scale automated legal writing evaluation.

%%
%% The next two lines define the bibliography style to be used, and
%% the bibliography file.
\bibliographystyle{ACM-Reference-Format}
\bibliography{references}

\end{document}